\documentclass[letterpaper]{article} % DO NOT CHANGE THIS
\usepackage{aaai2026}  % DO NOT CHANGE THIS
\usepackage{times}  % DO NOT CHANGE THIS
\usepackage{helvet}  % DO NOT CHANGE THIS
\usepackage{courier}  % DO NOT CHANGE THIS
\usepackage[hyphens]{url}  % DO NOT CHANGE THIS
\usepackage{amsfonts}
\usepackage{graphicx} % DO NOT CHANGE THIS
\usepackage{natbib}  % DO NOT CHANGE THIS AND DO NOT ADD ANY OPTIONS TO IT
\usepackage{caption} % DO NOT CHANGE THIS AND DO NOT ADD ANY OPTIONS TO IT
\usepackage{algorithm}
\usepackage{algorithmic}
\usepackage{amsmath}
\usepackage{multirow}
\usepackage{booktabs}
\usepackage{booktabs}
\usepackage{newfloat}
\usepackage{listings}
\DeclareCaptionStyle{ruled}{labelfont=normalfont,labelsep=colon,strut=off} % DO NOT CHANGE THIS
\floatstyle{ruled}
\newfloat{listing}{tb}{lst}{}
\floatname{listing}{Listing}
\title{Departures: Distributional Transport for Single-Cell Perturbation Prediction with Neural Schrödinger Bridges}
\author{
    Changxi Chi\textsuperscript{\rm 1,2}, Yufei Huang\textsuperscript{\rm 1,2}, Jun Xia\textsuperscript{\rm 3}, Jiangbin Zheng\textsuperscript{\rm 1,2}, Yunfan Liu\textsuperscript{\rm 1,2}, Zelin Zang\textsuperscript{\rm 4} and Stan Z. Li\textsuperscript{\rm 2}
}
\affiliations{
    \textsuperscript{\rm 1}Zhejiang University\\
    \textsuperscript{\rm 2}Westlake University\\
    \textsuperscript{\rm 3}The Hong Kong University of Science and Technology (Guangzhou)\\
    \textsuperscript{\rm 4}Centre for Artificial Intelligence and Robotics Hong Kong Institute of Science \& Innovation, Chinese Academy of Sciences\\
    
    \{12563057, huangyufei\}@zju.edu.cn
}

\usepackage{bibentry}
\begin{document}

\maketitle

\begin{abstract}
Predicting single-cell perturbation outcomes directly advances gene function analysis and facilitates drug candidate selection, making it a key driver of both basic and translational biomedical research. However, a major bottleneck in this task is the unpaired nature of single-cell data, as the same cell cannot be observed both before and after perturbation due to the destructive nature of sequencing. Although some neural generative transport models attempt to tackle unpaired single-cell perturbation data, they either lack explicit conditioning or depend on prior spaces for indirect distribution alignment, limiting precise perturbation modeling. In this work, we approximate Schrödinger Bridge (SB), which defines stochastic dynamic mappings recovering the entropy-regularized optimal transport (OT), to directly align the distributions of control and perturbed single-cell populations across different perturbation conditions. Unlike prior SB approximations that rely on bidirectional modeling to infer optimal source-target sample coupling, we leverage Minibatch-OT based pairing to avoid such bidirectional inference and the associated ill-posedness of defining the reverse process. This pairing directly guides bridge learning, yielding a scalable approximation to the SB. We approximate two SB models, one modeling discrete gene activation states and the other continuous expression distributions. Joint training enables accurate perturbation modeling and captures single-cell heterogeneity. Experiments on public genetic and drug perturbation datasets show that our model effectively captures heterogeneous single-cell responses and achieves state-of-the-art performance.

\end{abstract}

\begin{links}
    \link{Code}{https://github.com/ChangxiChi/Departures}
\end{links}

\section{Introduction}
Genetic perturbation and drug perturbation constitute the two primary avenues for studying and modeling single-cell responses to perturbations. Genetic perturbations, predominantly based on CRISPR technologies \cite{CRISPR_1,CRISPR_2}, target specific genes to reveal their functional roles, enabling systematic dissection of gene regulatory mechanisms at single-cell resolution. Drug perturbations, by contrast, modulate molecular pathways through small-molecule compounds \cite{scPerturb}, offering complementary insights into cellular responses and pharmacological effects. Together, these two perturbation strategies form the foundation of single-cell perturbation studies, providing mechanistic and translational perspectives on cell transitions under diverse experimental conditions. However, due to the high cost and limited scalability of single-cell experiments, it is impractical to measure all combinations of perturbations and cell types. This limitation highlights the need for computational models that can predict cellular responses under unseen perturbation conditions.\\
\begin{figure}[t]
    \centering
    \includegraphics[width=0.4\textwidth]{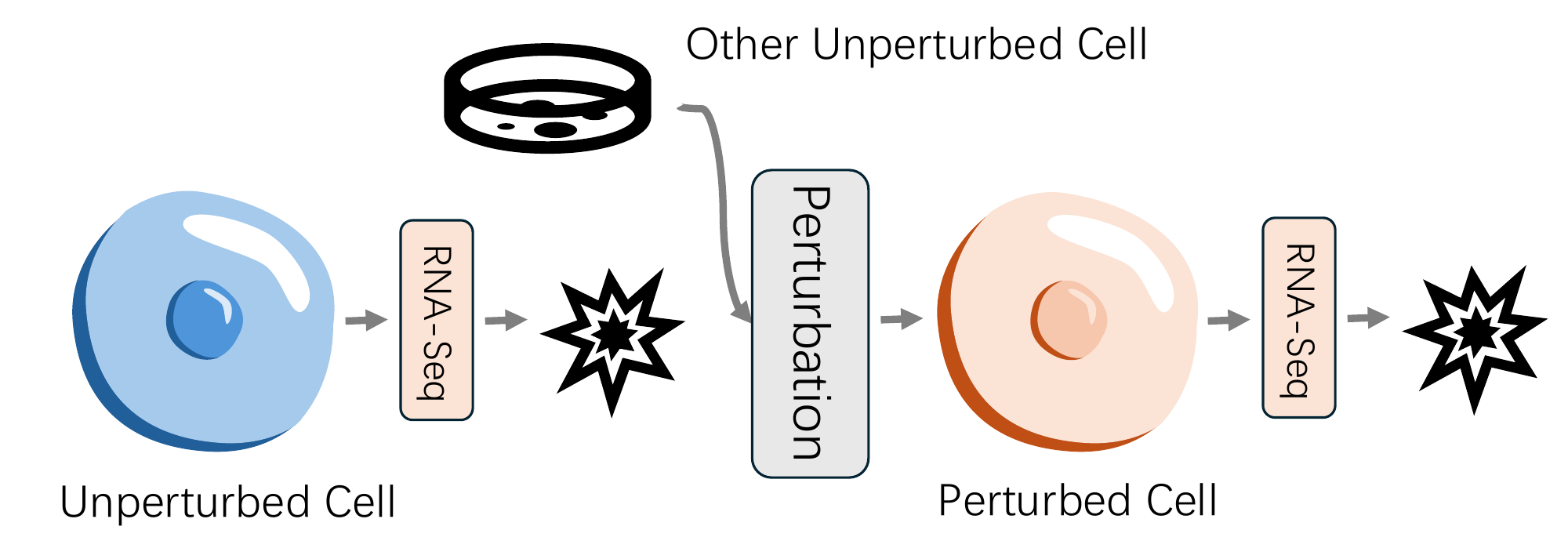}
    \caption{Single-cell perturbation data are unpaired as RNA-seq is destructive.}
    \label{fig:destructive}
\end{figure}
\indent A major challenge in this task is that single-cell perturbation data are inherently unpaired \cite{cellot,Unlasting}. This arises from the destructive nature of RNA sequencing, which requires cell lysis to release RNA content \cite{mortazavi2008mapping},  as illustrated in Fig.~\ref{fig:destructive}. As a result, it is impossible to observe the same cell under both control and perturbed conditions. Some studies \cite{cellot,scbutterfly} attempt to address this unpaired nature, but their use of unconditional models limits generalization to unseen perturbation conditions. In contrast, \cite{Unlasting} incorporates gene regulatory network (GRN) priors to model the effects of perturbations, providing a biologically grounded representation of cellular responses. It then aligns control and perturbed cell populations using dual conditional diffusion models with a shared prior space, enabling accurate prediction under unseen perturbation conditions. However, by aligning distributions indirectly through the prior space, it bypasses the underlying physical energy landscape, leading to transition paths that deviate from energy-optimal trajectories and fail to capture ideal optimal transport (OT).\\
\indent In this paper, we purpose \textbf{Departures} (\textbf{D}istributional Transport for Single-C\textbf{e}ll \textbf{P}erturb\textbf{a}tion P\textbf{r}edic\textbf{t}ion with Ne\textbf{u}ral Sch\textbf{r}öding\textbf{e}r Bridge\textbf{s}), a generative framework designed to predict cellular gene expression under diverse genetic and molecular perturbation conditions. \textbf{By directly aligning the distributions of control and perturbed cells, the framework captures population-level transitions without relying on potentially noisy explicit cell pairing or restrictive latent space alignment. This method reduces information loss, improves robustness to biological variability and noise, and provides a principled, scalable solution to the challenges posed by unpaired single-cell perturbation data.}\\
\indent While classical Schrödinger Bridge methods typically rely on alternating forward and backward updates \cite{IPF,DSBM} to iteratively refine the source-target node pairing between marginal distributions, the backward process is often ill-defined in conditional settings (Fig.~\ref{fig:indirect}) \cite{backward_2,backward}. Moreover, training two coupled models significantly increases computational cost and reduces scalability. To address this, we build upon Minibatch OT \cite{minibatch_OT,minibatch_OT_2} to directly enhance source-target sample coupling during training, without requiring alternating bidirectional model updates for pairing consistency. Built upon this alignment, we adopt bridge matching \cite{Markovian_proj_2,Markovian_proj_3} to learn the \textit{Markovian projection} \cite{Markovian_proj_1}, yielding a tractable approximation to the SB between control and perturbed distributions. To faithfully capture single-cell perturbation outcomes, we learn two SB models, the first models discrete gene activation states, and the second learns continuous expression dynamics. Joint training of these bridges enhances both biological fidelity and generative robustness.\\
\indent The main contributions of our work are as follows:
\begin{itemize}
\item We introduce \textbf{Departures}, a Schrödinger Bridge-based generative model that directly aligns the distributions of control and perturbed cells. To better approximate the Schrödinger Bridge, we use minibatch OT to compute source-target sample pairings that guide bridge matching training, eliminating the need for bidirectional iterative updates.
\item To model single-cell responses to perturbations, we design two Schrödinger bridge models, one for discrete gene activation and one for continuous expression, which are trained jointly to improve fidelity and robustness.
\item We demonstrate the superiority of \textbf{Departures} over existing methods on publicly available genetic and molecular perturbation datasets.
\end{itemize}

\section{Related Works and Preliminaries}
\subsection{Existing Perturbation Prediction Model}
There are numerous methods designed to predict single-cell perturbation responses, encompassing both genetic and drug-induced perturbations. These methods typically employ either generative models \cite{scGen,chemCPA,graphVCI,sams-vae,scgpt} or regression-based approaches \cite{GEARS,Squidiff,GRAPE}. However, many of these models overlook the inherently unpaired nature of single-cell perturbation data and fail to explicitly model the relationship between control and perturbed samples. While a few studies have addressed the unpaired nature of the data \cite{cellot,scbutterfly}, their reliance on unconditional models limits their ability to generalize to unseen perturbation settings. \cite{Unlasting} addresses this by incorporating gene regulatory network (GRN) priors to model perturbation effects, and aligning control and perturbed cell populations via dual conditional diffusion models that share a common prior space. Additionally, a regression-based mask model is introduced to predict gene expression status, thereby improving generation quality. However, this implicit alignment bypasses the true physical energy landscape, resulting in transitions that deviate from energy-optimal trajectories and fail to recover the ideal OT. Moreover, the mask model is trained separately, leading to suboptimal efficiency.

\begin{figure*}[t]
    \centering
    \includegraphics[width=\textwidth]{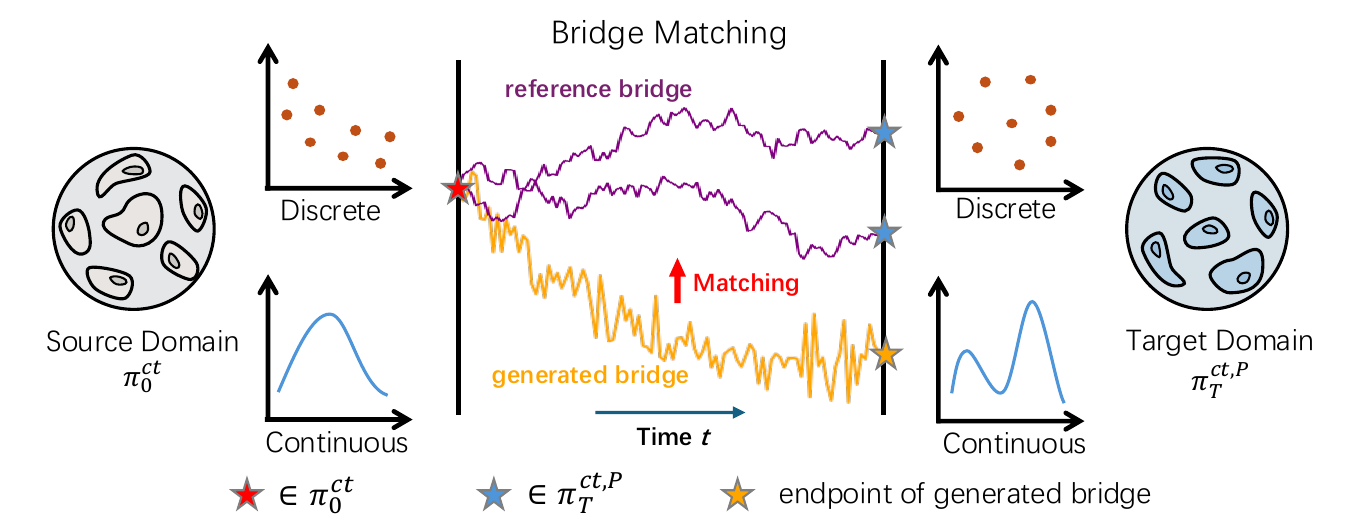}
    \caption{Overview of \textbf{Departures}. Conditioned on cell type $ct$, the model learns the distributional transition of single-cell gene expression profiles from the control population $\pi_{0}^{ct}$ to perturbed population $\pi_{T}^{ct,P}$. The problem is formulated by decoupling into two components: (1) modeling the distributional shift of gene expression levels before and after perturbation (\textbf{Continuous}), and (2) modeling the distributional shift of gene activation status induced by perturbation (\textbf{Discrete}).}
    \label{fig:overview}
\end{figure*}

\subsection{Schrödinger Bridges} 
The path measure space $\mathcal{P}(\mathcal{C})$ refers to the space of probability measures over continuous trajectories $x:[0,T]\to \mathbb{R}^d$, that is, $\mathcal{P}(\mathcal{C})=\mathcal{P}(C([0,T],\mathbb{R}^d))$. The Schrödinger Bridge (SB) problem \cite{SB_1} seeks a stochastic process $\mathbb{P}^{*}\in \mathcal{P}(\mathcal{C})$ that evolves between two given marginal distributions $\pi_{0}$ and $\pi_{T}$, while staying close to a reference process $\mathbb{Q} \in \mathcal{P}(\mathcal{C})$. It is formulated as:
\begin{equation}
    \mathbb{P}^{*}=\mathrm{argmin}_{\mathbb{P}\in \mathcal{P}(\mathcal{C})}\{ \mathrm{KL}(\mathbb{P}|\mathbb{Q}): \mathbb{P}_{0}=\pi_{0},\mathbb{P}_{T}=\pi_{T} \}
\end{equation}
The SB is an entropy-regularized dynamic Optimal Transport problem. Its path distribution implicitly defines an optimal pairing between nodes at the endpoints. \\
\indent A common approach to solving this problem is Iterative Proportional Fitting (IPF,\cite{IPF}), which alternates KL projections onto path measures with fixed initial or terminal marginals. However, it enforces marginal consistency without explicitly modeling the bridge dynamics.\\
\indent In contrast, following the formulation in \cite{DSBM}, the reference process $\mathbb{Q}$ is defined by a diffusion $\mathrm{d}\mathbf{X}_{t}=f_{t}(\mathbf{X}_{t})\mathrm{d}t + \sigma_{t}\mathrm{d}\mathbf{B}_{t}$, with $\mathbf{X}_{0}\in \mathbb{Q}_{0}=\pi_{0}$ and $(\mathbf{B}_{t})_{t\in[0,T]}$ is Brownian motion. To enforce $\mathbb{Q}_{T}$ also matches $\pi_{T}$, we construct a diffusion bridge $\mathbb{Q}_{|0,T}(\cdot|x_{0},x_{T})$ using Doob's $h$-transform \cite{Doob_h_t}, such that it is conditioned to start at $x_{0}$ and end at $x_{T}$:
\begin{equation}
    \mathrm{d}\mathbf{X}_{t}^{0,T}=\{ f_{t}(\mathbf{X}_{t}^{0,T}) +\sigma_{t}^{2} \nabla \mathrm{log}\mathbb{Q}_{T|t}(x_{T}|\mathbf{X}_{t}^{0,T}) \}\mathrm{d}t + \sigma_{t}\mathrm{d}\mathbf{B}_{t}
\end{equation}
where $\mathbf{X}_{0}^{0,T}=x_{0}$. Given $\Pi_{0,T}=\pi_{0}\otimes\pi_{T}$ and $\Pi=\Pi_{0,T}\mathbb{Q}_{|0,T}$, our goal is to find a Markov process $\mathrm{d}\mathbf{Y}_{t} = \{ f_{t}(\mathbf{Y}_{t}) + v_{t}(\mathbf{Y}_{t}) \} \mathrm{d}t + \sigma_{t}\mathrm{d}\mathbf{B}_{t}$. The implicit analytical solution of this problem is given by:
\begin{equation}
    v_{t}^{*}(x_{t})=\sigma_{t}^{2}\mathbb{E}_{\Pi_{T|t}}[\nabla\mathrm{log}\mathbb{Q}_{T|t}(\mathbf{X}_{T}|\mathbf{X}_{t})|\mathbf{X}_{t}=x_{t}]
    \label{analytical solution}
\end{equation}
which corresponds to the \textit{Markovian projection} \cite{Markovian_proj_1}. When $f_{t}=0$ and $\sigma_{t}=\sigma$, $\mathbb{Q}_{|0,T}$ is a Brownian Bridge, and we have: 
\begin{equation}
    \textbf{X}_{t}^{0,T}=\frac{t}{T}x_{T}+(1-\frac{t}{T})x_{0}+\sigma(\mathbf{B}_{t}-\frac{t}{T}\mathbf{B}_{T})
\label{eq:BuildBridge}
\end{equation}
\begin{equation}
    \mathrm{d}\mathbf{X}_{t}^{0,T}=\{ (x_{T}-\mathbf{X}_{t}^{0,T})/(T-t) \}\mathrm{d}t+\sigma\mathrm{d}\mathbf{B}_{t}
    \label{eq:BuildBridgeSDE}
\end{equation}
with $(\mathbf{B}_{t}-\frac{t}{T}\mathbf{B}_{T})\sim \mathcal{N}(0,t(1-\frac{t}{T})\mathrm{Id})$, and $\mathbf{B}_t$ is a standard Brownian motion.\\
\indent The objective of the model $v_{\theta}$ is to predict $v_{t}^{*}$. Specifically, the bridge matching \cite{Markovian_proj_3,Markovian_proj_2} operation learns the \textit{Markovian projection} by optimizing the following objective:
\begin{equation}
    \mathbb{E}_{\Pi_{t,T}}[|| (\mathbf{X}_{T} - \mathbf{X}_{t})/(T-t)-v_{\theta}(t,\mathbf{X}_{t}) ||^{2}]
    \label{v_pred}
\end{equation}
\indent To generate sample paths of the continuous bridge process, we implement the dynamics using the Euler–Maruyama method \cite{Euler–Maruyama}, a standard numerical scheme for approximating solutions to stochastic differential equations (SDEs).\\
\indent Based on the above, Iterative Markovian Fitting (IMF, \cite{DSBM}) is an algorithm designed to approximate Schrödinger bridges by alternating between \textit{Markovian projection} and \textit{Reciprocal projection} \cite{RP}. It employs two time-symmetric models to match the marginals at both ends of the path, iteratively refining their source-target sample coupling to approximate SB. However, learning both forward and backward models is challenging due to the inherently one-directional nature of perturbations.

\section{Methodology}
We introduce the proposed model \textbf{Departures} in this section. The overview is shown in Fig.~\ref{fig:overview}. Specifically, \textbf{Departures} learns the distributional transition mapping from control to perturbed samples, modeling both gene expression levels (\textbf{Continuous}) and gene activation states (\textbf{Discrete}) via bridge matching. To approximate the SB without requiring two time-reversal models, the node-level pairing is optimized using Minibatch OT \cite{minibatch_OT,minibatch_OT_2}.

\subsection{Problem Statement}
In single-cell perturbation prediction, the objective is to infer the gene expression profile of a specific cell type $ct$ under a given perturbation condition $P$. These conditions may stem from either genetic interventions or treatments with small-molecule compounds. For genetic perturbations, the condition is typically specified by the names of targeted genes, corresponding to gene knockout experiments. In contrast, small-molecule perturbations are characterized by the drug’s chemical structure along with its administered dosage. In our setting, instead of estimating a population-level summary (e.g., mean), the model generates a batch of samples under the given condition condition, aiming to represent the underlying distribution.

% \subsection{Data Preprocessing}
% The gene expression matrix is first log1p-normalized using SCANPY \cite{scanpy}, after which the top $N$ highly variable genes (HVGs) are selected.\\

\subsection{Distribution Transfer of Continuous Gene Expression}
Suppose under cell type $ct$ and perturbation condition $P$, we obtain a pairing $(x_{0},x_{T}) \sim \gamma,\gamma \in \Pi(\pi_{0}^{ct},\pi_{T}^{ct,P})$, where each $x_{0},x_{T}\in \mathbb{R}^{N}$ denotes a gene expression vector of a single cell, with $x_{0}$ sampled from the control distribution $\pi_{0}$, and $x_{T}$ sampled from the perturbed distribution $\pi_{T}^{P}$. A detailed discussion of the joint distribution $\gamma$ is provided later.\\
\indent To learn the transition between distributions, it is necessary to build bridges between $x_{0}$ and $x_{t}$. Specifically, we construct a diffusion bridge using the Doob's $h$-transform \cite{Doob_h_t}, following the assumption in \cite{Markovian_proj_2,DSBM} that $f_{t}=0$ and $\sigma_{t}=\sigma$. Referring to Eq.~\ref{eq:BuildBridge}, we have:
\begin{equation}
    x_{t}^{0,T}=\frac{t}{T}x_{T}+(1-\frac{t}{T})x_{0}+\sigma(\mathbf{B}_{t}-\frac{t}{T}\mathbf{B}_{T})
    \label{eq:interpolation}
\end{equation}
where $t\in[0,T]$, $\mathbf{B}_{t}\sim \mathrm{N}(0,\mathrm{Id})$, and $(\mathbf{B}_{t}-\frac{t}{T}\mathbf{B}_{T})\sim \mathrm{N}(0,t(1-\frac{t}{T})\mathrm{Id})$.\\
\indent The continuous bridge dynamics between the initial state and the fixed terminal point $x_T$ are described by the stochastic differential equation (SDE):
\begin{equation}
    \mathrm{d}x_{t}^{0,T}=
    v_{t}\mathrm{d}t+\sigma\mathrm{d}\mathbf{B}_{t},\quad \mathrm{d}\mathbf{B}_{t} \sim \mathcal{N}(0,\mathrm{Id})
    \label{eq:SDE1}
\end{equation}
where $v_{t}= (x_{T}-x_{t}^{0,T})/(T-t)$, $\sigma$ denotes noise scale, and $z \sim \mathcal{N}(0,1)$. Our objective is to train the model to learn the time-dependent drift term of this SDE, whose analytical solution is given by Eq.~\ref{analytical solution}, as guided by the $v_{t}$ prediction loss function in Eq.~\ref{v_pred}. In practice, sample paths of this SDE are generated by discretizing it using the Euler–Maruyama method \cite{Euler–Maruyama}.\\
\indent Rather than regressing the velocity field $v_t$ directly, we instead predict the endpoint $x_T$, which can be equivalently transformed to the drift $v_t$. This alternative formulation leads to more stable optimization and reduced numerical error \cite{x_T_pred}.\\
\indent Moreover, due to the sparsity of gene expression data, directly computing the loss over all genes can lead to mode collapse \cite{Unlasting}. To address this, we compute the loss only over genes with non-zero expression after perturbation. A separate module for predicting gene expression status will be introduced in a later section. As a result, the model $x_{\theta}$ focuses on expressed genes during training. The final objective function is defined as follows:
\begin{equation}
    \mathcal{L}_{cont}=
    \mathbb{E}_{t,(x_{0},x_{T})\sim\gamma}
    \left[ \frac{||d_{T}\odot (x_{T}-x_{\theta}(t,x_{t}^{0,T},ct,P)) ||^{2}}{\sum_i d_{T,i}}
    \right]
    \label{end_pred}
\end{equation}
where $\odot$ denotes Hadamard Product, $d_{T}\in \{0,1\}^{N}$ is a binary mask vector defined as:
\begin{equation}
    d_{T,i} = \begin{cases}
        1, & \text{if } x_{T,i} \neq 0 \\
        0, & \text{otherwise}
    \end{cases}
    \label{eq:transfer_into_discrete}
\end{equation}

\subsection{Distribution Transfer of Discrete Gene Status}
As previously mentioned, gene expression data are high-dimensional and sparse. To prevent the model from treating all genes equally and thus causing mode collapse, we apply a mask to the loss in Eq.~\ref{end_pred} based on the true gene expression status at the endpoint. Therefore, it is necessary to train a separate module capable of accurately predicting whether a gene is expressed or not. \\
\indent Given a pairing $(x_{0},x_{T})\sim\gamma$, we convert it into discrete labels $(d_{0},d_{T})\sim\gamma_{d}$ using Eq.~\ref{eq:transfer_into_discrete}. However, in the discrete setting, directly applying the interpolation in Eq.\ref{eq:interpolation} to the discrete labels is not meaningful, as the labels represent categorical states rather than continuous values. We define the intermediate states of the bridge process as stochastic mixtures of the two endpoints according to the following distribution:
\begin{equation}
    d_{t,i}=\begin{cases}
        x_{T,i} &\text{with prob} = \kappa(t)\\
        x_{0,i} &\text{with prob} = (1-\kappa(t))
    \end{cases}
\end{equation}
where we set $\kappa(t)=\frac{t}{T}$ is monotonic increasing with $\kappa(0)=0$ and $\kappa(T)=1$. This process captures the discrete interpolation from the initial state to the terminal state by randomly selecting each coordinate from $x_{0}$ or $x_{T}$according to $\kappa(t)$. \\
\indent Unlike diffusion bridges in continuous spaces governed by SDE, the discrete bridge corresponds to a continuous-time Markov chains (CTMC, \cite{CTMC_1,CTMC_2}) on discrete space characterized by the transition probability:
\begin{equation}
    p_{t+h|t}(d_{t+h}|d_{t})=\delta(d_{t+h},d_{t})+hu_{t}(d_{t+h},d_{t})+o(h)
    \label{eq:CTMC}
\end{equation}
where $d_{t},d_{t+h}\in \{0,1\}^{N}$, $\delta$ is Kronecker delta function, and $u_{t}(d_{t+h},d_{t})$ represents transition rate from $d_{t}$ to $d_{t+h}$. Our goal is to learn the transition rate function $u_t$, which fully characterizes the dynamics of the CTMC.\\
\indent As shown in \cite{DFM}, this can be achieved by learning the conditional distribution $p_{T|t}$ as:
\begin{align}
    u_{t}(d_{t+h}, d_{t}) = \mathbb{E}_{d_{T}} \Bigg[ 
    & \frac{\kappa(t)}{1-\kappa(t)} \big[ \delta(d_{t+h}, d_T) - \delta(d_{t+h}, d_t) \big] \nonumber \\
    & \quad \times p_{T|t}(d_T | d_t) 
    \Bigg]
    \label{eq:marginal}
\end{align}
with $p_{T|t}(d_T | d_t)=\mathbb{E}_{(d_{0},d_{T})\sim\gamma_{d}}p_{0,1|t}(d_{0},d_{T}|d_{t})$. We aim to approximate this conditional posterior $p_{T|t}(\cdot|d_{t})$ using a parameterized neural network.\\
\indent Intuitively, learning the posterior transition probabilities $p_{T|t}(\cdot|d_{t})$ is correspond to the conditional distribution of the bridge’s terminal state given the intermediate state. The objective function is as follows:
\begin{equation}
    \mathcal{L}_{disc}=\mathbb{E}_{t,(d_{0},d_{T})\sim\gamma_{d}}\left[ D(d_{T},d_{\theta}(t,d_{t},ct,P)) \right]
    \label{disc_loss}
\end{equation}
where \(D(\cdot, \cdot)\) denotes the cross-entropy loss between the true discrete state $d_{T}$ and the prediction.

\begin{figure}[t]
    \centering
    \includegraphics[width=0.45\textwidth]{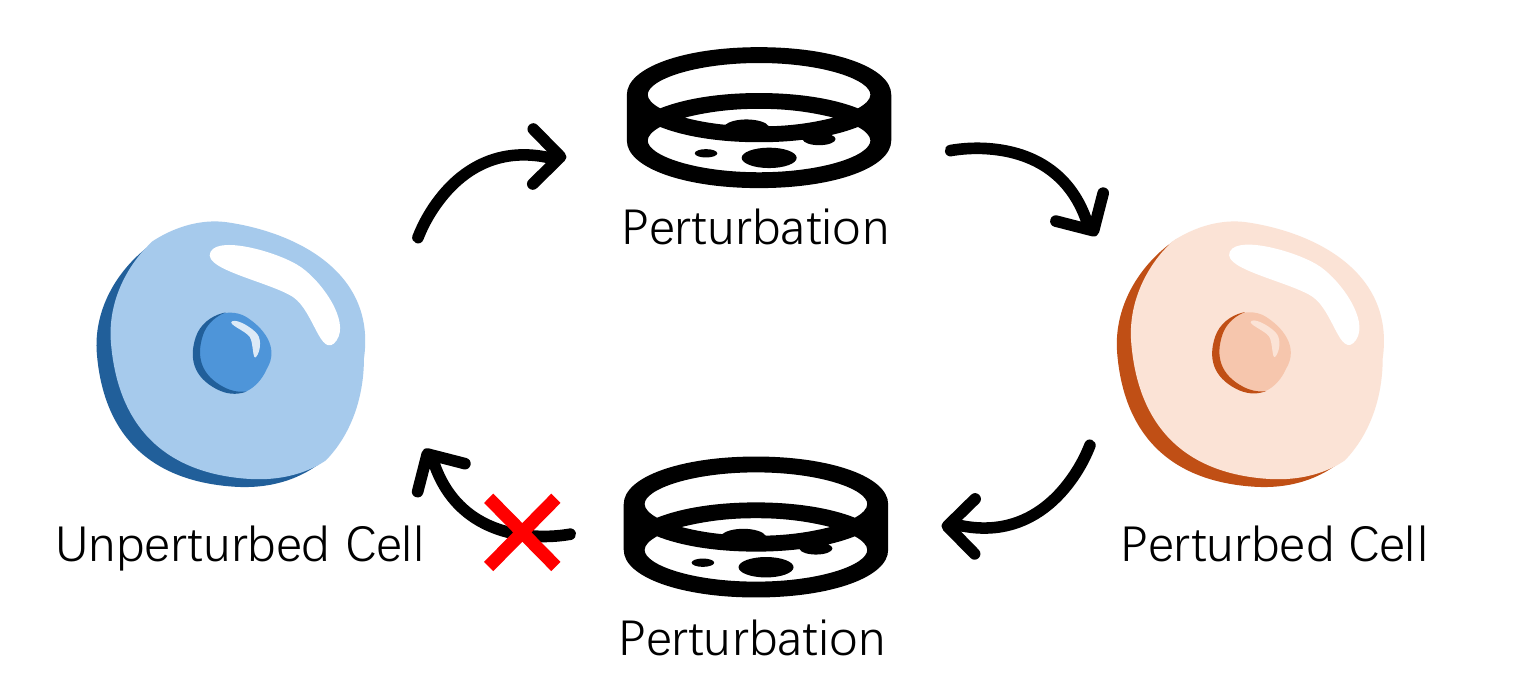}
    \caption{Perturbations induce natural, directional transitions from control to treated states. In contrast, reconstructing control states from perturbed ones is ill-posed, and training a backward model adds significant computational overhead.}
    \label{fig:indirect}
\end{figure}

\subsection{Initialization Pairing for Schrödinger Bridge}
We now discuss the joint distribution $\gamma$. As introduced earlier, IMF\cite{DSBM} employs two time-symmetric neural networks to model the forward and backward processes, matching the marginal distributions at both ends of the trajectory through Markovian projection and reciprocal projection. A core design goal of IMF is to reduce error accumulation during node rectification, where the joint distribution $\gamma$ is iteratively updated. By obtaining higher-quality pairing between nodes from $\gamma$, it subsequently facilitates the learning of the diffusion bridge. Through iterative optimization, the process converges to the true SB.\\
\indent However, a major obstacle in applying IMF to conditional generation tasks is the need to learn models in both forward and backward directions. The backward process of generating from the perturbed distribution back to the control distribution is ill-defined (Fig.~\ref{fig:indirect}), which makes bidirectional training impractical \cite{backward_2,backward}.\\
\indent To address this problem, we draw inspiration from Minibatch OT \cite{minibatch_OT,minibatch_OT_2} to construct local OT pairings by selecting subsets of control samples for all samples under each perturbation condition. Specifically, under perturbation conditon $P$ and cell type $ct$, we obtain a batch of perturbed cell samples $\{x_{T}^{(i)}\}^{B}_{i=1}$. Correspondingly, we sample an equal number of control samples $\{x_{0}^{(i)}\}^{B}_{i=1}$ from the same cell type $ct$ to serve as the source for pairing. Then we compute an optimal transport (OT) plan $\pi_{batch}$ between these two sets of samples using the Sinkhorn algorithm \cite{sinkhorn}, which efficiently solves the entropically regularized OT problem. The OT plan defines a joint distribution over matched pairs $(x_{0}^{(j)}, x_{T}^{(i)})$ that approximately minimizes the transport cost within each batch. Based on these pairs, we apply Eq.~\ref{eq:transfer_into_discrete} to derive the corresponding discrete gene activation states $(d_{0}^{(j)}, d_{T}^{(i)})$. The continuous and discrete pairings define the endpoints of the bridge, guiding the model to learn \textit{Markovian projections} via bridge matching (Eq.~\ref{end_pred}, Eq.~\ref{disc_loss}), thereby approximating the Schrödinger Bridge.

\subsection{Implementation and Generation}
Before training, we perform log1p normalization on the gene expression matrix using Scanpy \cite{scanpy}, followed by the selection of the top $N$ highly variable genes (HVGs). In modeling various perturbation conditons, we adopt the approach of \cite{Unlasting}, which incorporates the Gene Regulation Network (GRN) to model perturbation conditions. During training, we initialize the source-target sample couplings at the beginning of each epoch using the strategy described in the \textbf{for Schrödinger bridge Section}. We jointly train the model $x_{\theta}$ predicting changes in continuous gene expression values and the model $d_{\theta}$ predicting changes in discrete gene activation status introduced above, with the following loss function:
\begin{equation}
    \mathcal{L}=\mathcal{L}_{cont}+\mathcal{L}_{disc}
\end{equation}
\indent In predicting the perturbation results, we perform sampling separately from the two models to obtain predictions of gene expression values $\hat{x}_{T}$ and gene activation states $\hat{d}_{T}$. Specifically, when predicting perturbation outcomes of cells of type $ct$ under perturbation condition $P$, we iteratively generate the results according to the following procedure, as introduced in Eq.~\ref{eq:SDE1}:
\begin{equation}
    \mathrm{d}x_{t}=
    \frac{(x_{\theta}(t,x_{t},ct,P)-x_{t})}{(T-t)}\mathrm{d}t+\sigma\mathrm{d}\mathbf{B}_{t}
    \label{eq:gen1}
\end{equation}
where at time $t=0$, $x_{0}$ denotes a sample drawn from the control group of the same cell type $ct$, and at time $t=T$, $x_{T}=\hat{x}_{T}$ represents the final predicted outcome.\\
\indent When predicting the discrete activation states of genes, we follow \cite{DFM} and define the conditional transition probability of moving from state $d_{t}$ at time $t$ to the next state after a small increment $h$ as follow:
\begin{equation}
    p_{t+h|t}(\cdot|d_{t})=d_{t}+h\times\frac{d_{\theta}(t,d_{t},ct,P)}{T-t}
    \label{eq:gen2}
\end{equation}
where $d_{0}$ denotes the discrete initial state representing gene activation of a control sample of cell type $ct$. The next state $d_{t+h}$ is then sampled from the distribution $p_{t+h|t}(\cdot|d_{t})$, and by iteratively applying this sampling procedure over time steps, the final predicted discrete activation state $d_{T}=\hat{d}_{T}$ is obtained.\\
\indent In practice, to generate predictions, we first sample a real control group example $x_{0}$ from cell $ct$ and its corresponding discretized state $d_{0}$ (Eq.~\ref{eq:transfer_into_discrete}) as the respective starting points for the continuous and discrete models. We then perform iterative sampling using Eq.~\ref{eq:gen1} and Eq.~\ref{eq:gen2} to obtain $\hat{x}_{T}$ and $\hat{d}_{T}$. The final predicted outcome $\tilde{x}_{T}$ is given by:
\begin{equation}
   \tilde{x}_{T} =\hat{x}_{T}\odot \hat{d}_{T}
\end{equation}
where $\odot$ denotes Hadamard Product.

\begin{table*}[]
\small
\centering
\resizebox{0.85\textwidth}{!}{
\begin{tabular}{cccccccc}
\hline
\multirow{2}{*}{\raisebox{-0.5\height}{\textbf{Dataset}}} 
& \multirow{2}{*}{\raisebox{-0.5\height}{\textbf{Methods}}} 
& \multicolumn{2}{c}{All}                                                                                    
& \multicolumn{2}{c}{DE20}                                                                                   
& \multicolumn{2}{c}{DE40}                                                                                          
\\ \cmidrule(lr){3-4} \cmidrule(lr){5-6} \cmidrule(lr){7-8}
\multicolumn{2}{c}{}                   
& E-distance($\downarrow$)                                              
& EMD($\downarrow$)                                                     
& E-distance($\downarrow$)                                              
& EMD($\downarrow$)                                                     
& E-distance($\downarrow$)                                              
& EMD($\downarrow$)                                                     \\ \hline
\multirow{10}{*}{Adamson}   
                           & GRAPE %\cite{GRAPE}     
                           & \begin{tabular}[c]{@{}c@{}}0.7905\\ $\pm$0.0484\end{tabular} 
                           & \begin{tabular}[c]{@{}c@{}}0.0444\\ $\pm$0.0024\end{tabular} 
                           & \begin{tabular}[c]{@{}c@{}}0.7514\\ $\pm$0.0575\end{tabular} 
                           & \begin{tabular}[c]{@{}c@{}}0.1528\\ $\pm$0.0234\end{tabular} 
                           & \begin{tabular}[c]{@{}c@{}}0.7648\\ $\pm$0.0565\end{tabular} 
                           & \begin{tabular}[c]{@{}c@{}}0.1503\\ $\pm$0.0182\end{tabular} \\
                           \cmidrule(lr){3-8}
                           & GEARS %\cite{GEARS}
                           & \begin{tabular}[c]{@{}c@{}}0.8721\\ $\pm$0.1304\end{tabular} 
                           & \begin{tabular}[c]{@{}c@{}}0.0531\\ $\pm$0.0027\end{tabular} 
                           & \begin{tabular}[c]{@{}c@{}}0.7884\\ $\pm$0.1245\end{tabular} 
                           & \begin{tabular}[c]{@{}c@{}}0.1298\\ $\pm$0.0324\end{tabular} 
                           & \begin{tabular}[c]{@{}c@{}}0.7935\\ $\pm$0.0544\end{tabular} 
                           & \begin{tabular}[c]{@{}c@{}}0.1221\\ $\pm$0.0231\end{tabular} \\
                           \cmidrule(lr){3-8}
                           & GraphVCI %\cite{graphVCI}
                           & \begin{tabular}[c]{@{}c@{}}1.3182\\ $\pm$0.9763\end{tabular} 
                           & \begin{tabular}[c]{@{}c@{}}0.3026\\ $\pm$0.1953\end{tabular} 
                           & \begin{tabular}[c]{@{}c@{}}2.4499\\ $\pm$0.2446\end{tabular} 
                           & \begin{tabular}[c]{@{}c@{}}1.2457\\ $\pm$0.5183\end{tabular} 
                           & \begin{tabular}[c]{@{}c@{}}2.6327\\ $\pm$0.4950\end{tabular} 
                           & \begin{tabular}[c]{@{}c@{}}1.0801\\ $\pm$0.0866\end{tabular} \\
                           \cmidrule(lr){3-8}
                           & scGPT %\cite{scgpt}   
                           & \begin{tabular}[c]{@{}c@{}}2.1368\\ $\pm$0.0441\end{tabular} 
                           & \begin{tabular}[c]{@{}c@{}}0.1724\\ $\pm$0.0355\end{tabular} 
                           & \begin{tabular}[c]{@{}c@{}}1.2571\\ $\pm$0.3373\end{tabular} 
                           & \begin{tabular}[c]{@{}c@{}}0.3895\\ $\pm$0.1032\end{tabular} 
                           & \begin{tabular}[c]{@{}c@{}}1.4484\\ $\pm$0.3087\end{tabular} 
                           & \begin{tabular}[c]{@{}c@{}}0.3781\\ $\pm$0.0866\end{tabular} \\ 
                            \cmidrule(lr){3-8}
                           & \textbf{Departures} (Ours)
                            & \begin{tabular}[c]{@{}c@{}}\textbf{0.5955}\\ $\pm$0.1212\end{tabular} 
                            & \begin{tabular}[c]{@{}c@{}}\textbf{0.0356}\\ $\pm$0.0053\end{tabular} 
                            & \begin{tabular}[c]{@{}c@{}}\textbf{0.5028}\\ $\pm$0.1942\end{tabular} 
                            & \begin{tabular}[c]{@{}c@{}}\textbf{0.0953}\\ $\pm$0.0487\end{tabular} 
                            & \begin{tabular}[c]{@{}c@{}}\textbf{0.5243}\\ $\pm$0.2243\end{tabular} 
                            & \begin{tabular}[c]{@{}c@{}}\textbf{0.1013}\\ $\pm$0.0442\end{tabular} \\
                           \hline
\multirow{8}{*}{sci-Plex3} 
                           & chemCPA %\cite{chemCPA}
                           & \begin{tabular}[c]{@{}c@{}}0.7847\\ $\pm$0.1029\end{tabular} 
                           & \begin{tabular}[c]{@{}c@{}}0.0838\\ $\pm$0.0081\end{tabular} 
                           & \begin{tabular}[c]{@{}c@{}}0.4717\\ $\pm$0.1571\end{tabular} 
                           & \begin{tabular}[c]{@{}c@{}}0.1836\\ $\pm$0.0358\end{tabular} 
                           & \begin{tabular}[c]{@{}c@{}}0.5008\\ $\pm$0.1659\end{tabular} 
                           & \begin{tabular}[c]{@{}c@{}}0.1784\\ $\pm$0.0261\end{tabular} \\
                           \cmidrule(lr){3-8}
                           & CPA %\cite{cpa}   
                           & \begin{tabular}[c]{@{}c@{}}0.9894\\ $\pm$0.1336\end{tabular}            
                           & \begin{tabular}[c]{@{}c@{}}0.1357\\ $\pm$0.0461\end{tabular}  
                           & \begin{tabular}[c]{@{}c@{}}0.9737\\ $\pm$0.9768\end{tabular} 
                           & \begin{tabular}[c]{@{}c@{}}0.3761\\ $\pm$0.0667\end{tabular} 
                           & \begin{tabular}[c]{@{}c@{}}1.0794\\ $\pm$1.1890\end{tabular} 
                           & \begin{tabular}[c]{@{}c@{}}0.3856\\ $\pm$0.0387\end{tabular} \\
                           \cmidrule(lr){3-8}
                           & GraphVCI %\cite{graphVCI}
                           & \begin{tabular}[c]{@{}c@{}}0.8393\\ $\pm$0.1823\end{tabular} 
                           & \begin{tabular}[c]{@{}c@{}}0.0986\\ $\pm$0.0108\end{tabular} 
                           & \begin{tabular}[c]{@{}c@{}}0.4958\\ $\pm$0.1275\end{tabular} 
                           & \begin{tabular}[c]{@{}c@{}}0.2016\\ $\pm$0.0379\end{tabular} 
                           & \begin{tabular}[c]{@{}c@{}}0.5174\\ $\pm$0.1347\end{tabular} 
                           & \begin{tabular}[c]{@{}c@{}}0.1861\\ $\pm$0.0288\end{tabular} \\
                           \cmidrule(lr){3-8}
                           & \textbf{Departures} (Ours)
                            & \begin{tabular}[c]{@{}c@{}}\textbf{0.5478}\\ $\pm$0.1461\end{tabular} 
                            & \begin{tabular}[c]{@{}c@{}}\textbf{0.0254}\\ $\pm$0.0051\end{tabular} 
                            & \begin{tabular}[c]{@{}c@{}}\textbf{0.2764}\\ $\pm$0.1246\end{tabular} 
                            & \begin{tabular}[c]{@{}c@{}}\textbf{0.0623}\\ $\pm$0.0317\end{tabular} 
                            & \begin{tabular}[c]{@{}c@{}}\textbf{0.2827}\\ $\pm$0.1113\end{tabular} 
                            & \begin{tabular}[c]{@{}c@{}}\textbf{0.0565}\\ $\pm$0.0340\end{tabular} \\                   
                           \hline
\end{tabular}}
\caption{Performance comparison on Adamson and sci-Plex3 datasets, evaluated using E-distance and $\text{EMD}$ on all genes, top 20, and top 40 differentially expressed (DE) genes.}
\label{tab:performance_comparison}
\end{table*}

\section{Experiments and Results}
\subsection{Dataset}
In the main experiments, we leverage diverse datasets covering both genetic and chemical perturbations to evaluate our method, including the Adamson CRISPR knockout dataset \cite{adamson} and the sci-Plex3 chemical perturbation dataset \cite{sciplex3}. The Adamson dataset comprises gene expression profiles from 87 distinct single-gene perturbations in a single cell type, while the sci-Plex3 dataset spans 187 chemical perturbations across four dosage levels and three different cell types. Each perturbation condition in both datasets is represented by the average profile of hundreds of single cells, ensuring robust statistical coverage. We consider 5,000 genes in the Adamson dataset and 2,000 genes in sci-Plex3, providing a rich and representative foundation for model training and evaluation. By evaluating our model across different perturbation modalities and biological settings, we demonstrate its broad applicability and effectiveness.
\subsection{Experiment Settings}
We conduct experiments on two widely-used datasets in the single-cell perturbation modeling field, Adamson \cite{adamson} and SciPlex3 \cite{sciplex3}. For the Adamson dataset, we randomly split the perturbed genes into 70\% for training and 30\% for testing across all gene perturbation types. All control group cells are included in the training set. For the SciPlex3 dataset, all control group cells are included in the training set. For each experimental condition (defined by a unique combination of drug, dosage, and cell type), we assign the entire group of corresponding cells to the test set with a 30\% probability; otherwise, it is assigned to the training set. We select the top 5,000 highly variable genes (HVGs) for the Adamson dataset and the top 2,000 HVGs for the SciPlex3 dataset. All baseline methods are evaluated using the same gene selection to ensure a fair comparison.\\
\indent The model is trained using the AdamW \cite{AdamW} optimizer with a learning rate of 0.001 and a batch size of 64. During inference, we use 50 uniformly spaced time steps in Eq.~\ref{eq:gen1} and Eq.~\ref{eq:gen2} (i.e., a step size of 0.02). We set the noise scale $\sigma$ in Eq.~\ref{eq:gen1} to 0.2. All our method and its competitors are conducted using two Nvidia A100 GPU.\\
\indent For evaluation, we follow previous work and adopt distribution-aware metrics to account for the strong heterogeneity observed in single-cell data. Specifically, we use Energy Distance (E-distance) to capture overall distributional alignment by considering both inter-group and intra-group distances, and Earth Mover’s squared Distance ($\text{EMD}$) to quantify gene-level shifts by measuring the minimal cost to align predicted and true distributions. These metrics together provide a comprehensive and robust assessment of model performance at both the population and gene levels. Detailed computation procedures can be found in the Appendix.

\begin{figure}[t]
    \centering
    \includegraphics[width=0.45\textwidth]{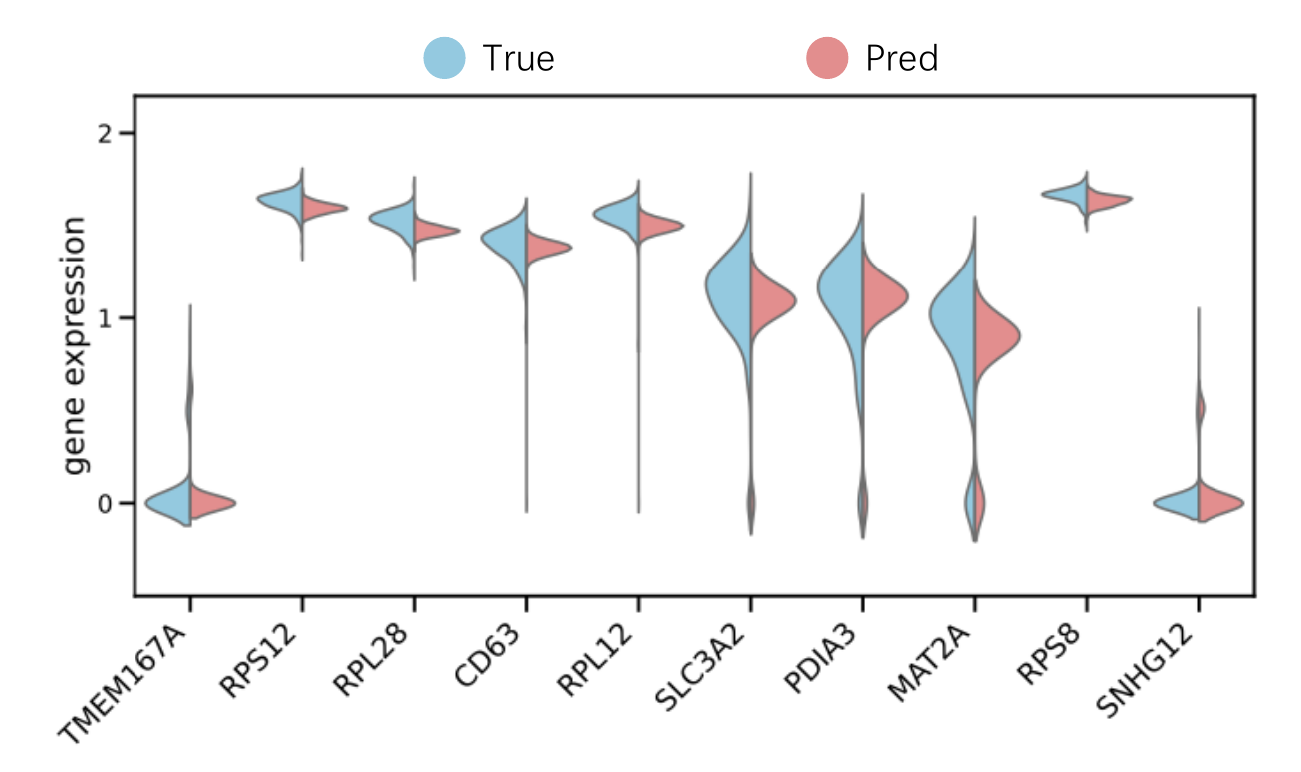}
    \caption{Violin plots comparing predicted and actual expression levels of the top differentially expressed (DE) genes under the TMEM167A knockout condition, which was unseen during training, from the Adamson dataset.}
    \label{fig:case}
\end{figure}

\subsection{\textbf{Departures} Outperform Existing Methods}
To assess the effectiveness of our model in predicting gene expression under perturbations, we compare it against several baseline methods, including GraphVCI \cite{graphVCI}, scGPT \cite{scgpt}, CPA \cite{cpa}, chemCPA \cite{chemCPA}, GEARS \cite{GEARS}, and GRAPE \cite{GRAPE}. For regression-based methods (GEARS and GRAPE), we augment their outputs by adding the predicted gene expression changes to different control samples. This promotes diversity in the generated profiles and ensures a more equitable comparison.\\
\indent Table \ref{tab:performance_comparison} shows that \textbf{Departures} outperforms graphVCI, scGPT, GEARS, GRAPE, which rely on forced pairing of perturbed and unperturbed cells during training. This forced matching can limit the model's ability to capture true biological variability and cell heterogeneity. Although GEARS and GRAPE inherit control group variability by adding a fixed predicted delta to control samples, this approach does not fully translate the control group's diversity into the predicted results. Because they do not condition their predictions on the starting point information, their performance is inferior to that of Departures.\\
\indent Methods such as \cite{chemCPA} and \cite{cpa} focus solely on reconstructing perturbed cells without modeling the transitional dynamics from unperturbed states. By relying only on predictions of mean and variance, these approaches oversimplify the perturbation process, limiting their ability to capture the inherent complexity and heterogeneity of cellular responses. In contrast, our model starts from the initial state and guides the state transitions through the learned Schrödinger bridge dynamics. Moreover, as illustrated in Fig.~\ref{fig:case}, our model accurately captures the distribution of top differentially expressed (DE) genes after perturbation.

\begin{figure}[t]
    \centering
    \includegraphics[width=0.4\textwidth]{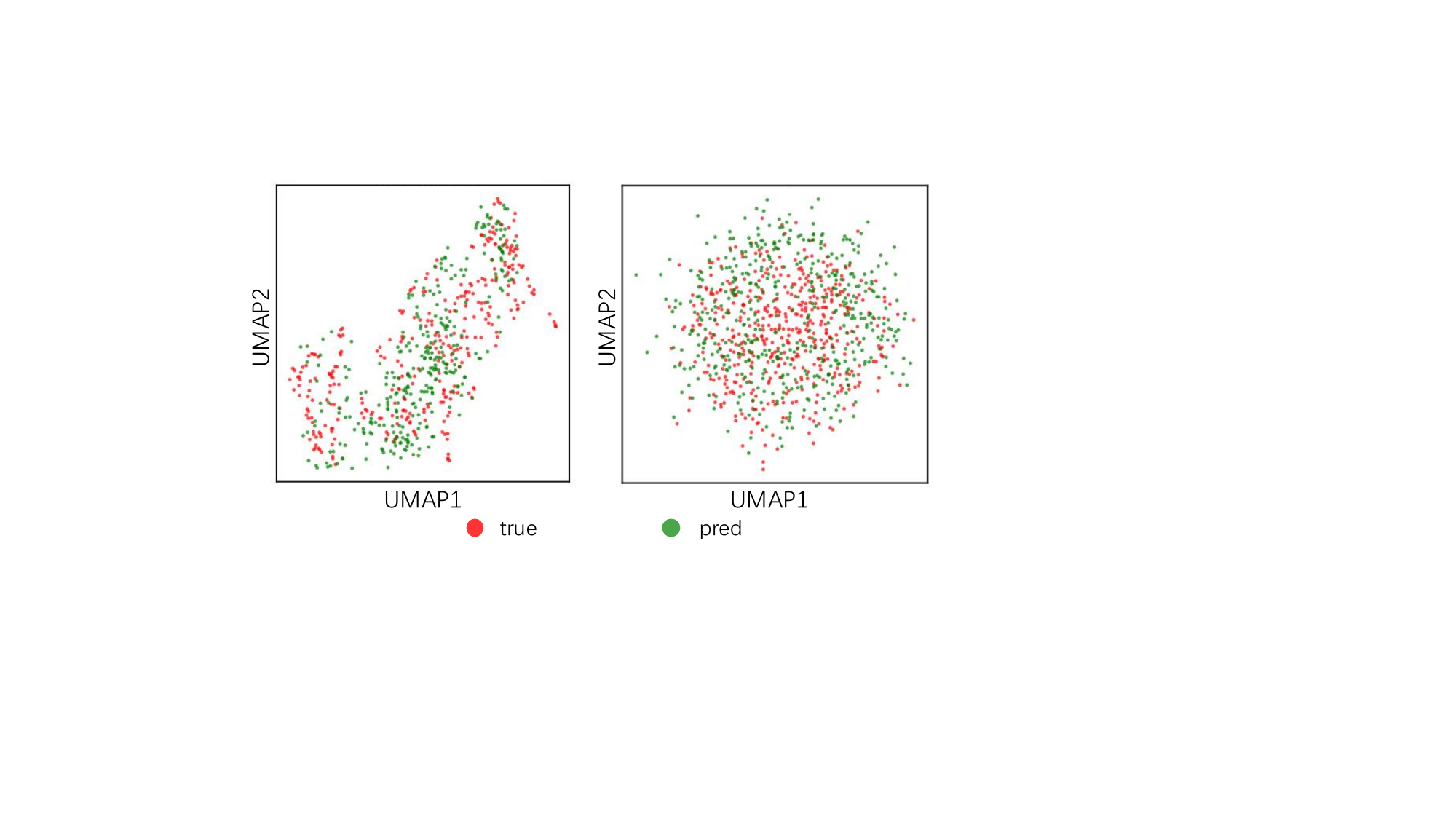}
    \caption{UMAP visualization of predicted and actual gene activation states. The left panel shows results for the unseen HSD17B12 perturbation condition from the Adamson test set. The right panel presents predictions for a held-out perturbation (compound Sodium – dosage 0.001 – cell type MCF7) from the sciplex3 test set.}
    \label{fig:umap}
\end{figure}

\begin{table}[ht]
\centering
\small
\begin{tabular}{ccc}
\hline
          & Adamson & sciplex3 \\ \hline
PCC\_all  & 0.9910$\pm$0.0078     & 0.9652$\pm$0.0690      \\ \hline
PCC\_DE20 & 0.9515$\pm$0.0685     & 0.9437$\pm$0.0969      \\ \hline
PCC\_DE40 & 0.9493$\pm$0.0716     & 0.9520$\pm$0.0722      \\ \hline
\end{tabular}
\caption{Pearson correlation coefficients (PCC) between the predicted gene activation probabilities after perturbation and the true probabilities.}
\label{tab:PCC}
\end{table}

\subsection{Discrete Model Effectively Captures Gene Activation States}
To evaluate the effectiveness of our discrete bridge model in predicting gene activation states after perturbation, we showcase the Pearson correlation coefficients (PCC) between predicted and true gene activation probabilities in Table~\ref{tab:PCC}. Specifically, under each perturbation condition, we first predict the activation states of $k$ cells $\{\hat{d}_{T}^{(i)}\}^{k}\in \{0,1\}^{k\times N}$ using Eq.~\ref{eq:gen2}, and then compute gene-wise activation probabilities. Similarly, we estimate the ground-truth activation probabilities from the real cell samples under the same condition, and compute the PCC between predicted and true probabilities. Additionally, we visualize the predicted gene activation states of individual cells in Fig.~\ref{fig:umap}. The high PCCs and visualization results show strong alignment between predicted and actual gene activation states.

\subsection{Ablation Study}
\begin{figure}[t]
    \centering
    \includegraphics[width=0.4\textwidth]{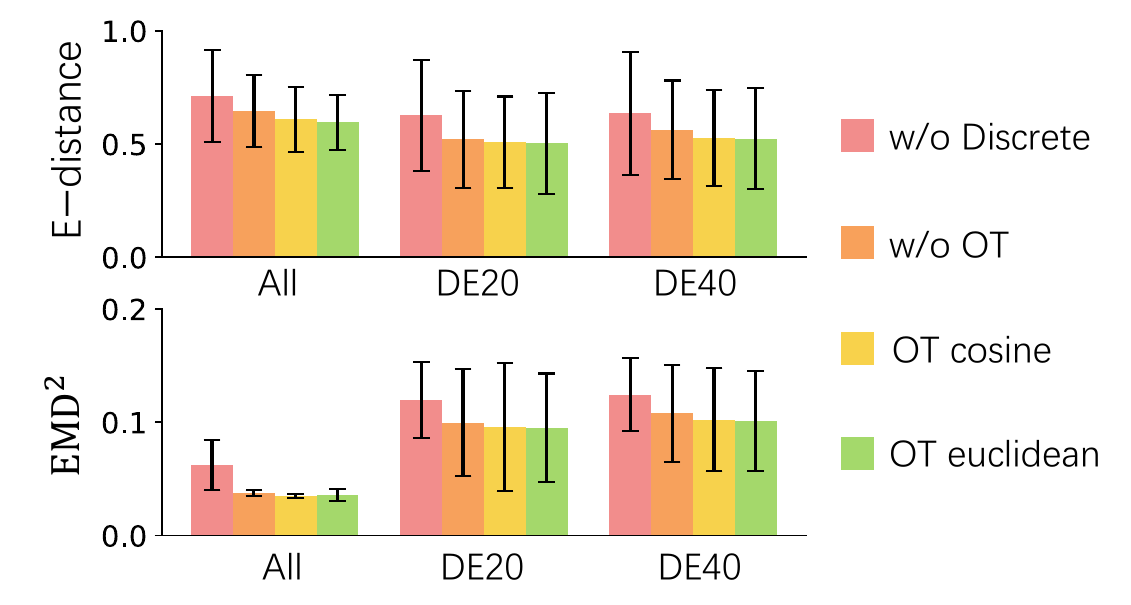}
    \caption{Ablation study results on Adamson.}
    \label{fig:ablation}
\end{figure}

To further evaluate the effectiveness of \textbf{Departures}, we compare it with the following methods through an ablation study. 1)\textbf{w/o~Discrete}: Only uses a single model to capture continuous gene expression, without discrete discrete activation prediction model. 2)\textbf{w/o~OT}: Node pairing is performed without Minibatch-OT based pairing, relying on random pairing instead. 3)\textbf{OT~cosine}: Uses cosine similarity as the transport cost when computing the OT-based pairing between unperturbed and perturbed distributions. 4)\textbf{OT~euclidean}: Uses Euclidean distance as the transport cost when computing the OT-based pairing between unperturbed and perturbed distributions. The results are shown in Fig.\ref{fig:ablation}.\\
\indent Experimental results show that modeling gene activation with a discrete model plays a crucial role in accurately predicting gene expression. It helps the prediction focus on important genes rather than overfitting sparse regions, thereby preventing mode collapse. Besides, OT-based pairing, inspired by the Schrödinger Bridge, outperforms random pairing by learning a probabilistically consistent transport plan that better captures biological transitions.

\section{Conclusion}
In this work, we present \textbf{Departures}, a generative framework for distribution-level prediction of single-cell responses under various perturbation conditions, via an approximation of the Schrödinger Bridge. Considering that the reverse process of conditional generation under gene perturbation is ill-posed, we employ Minibatch-OT to obtain a relatively better source-target sample coupling. This approach avoids the need to learn bidirectional conditional models with iterative pairing updates as before. Based on this pairing, we learn two \textit{Markovian projections} through bridge matching, one capturing discrete gene activation patterns and the other modeling continuous expression changes. Together, these enhance biological fidelity and generative robustness, enabling an efficient approximation of the Schrödinger Bridge. This approach learns distributional transitions between pre- and post-perturbation cells, effectively addressing the unpaired nature of single-cell perturbation data. By integrating gene regulatory priors and jointly modeling discrete and continuous dynamics, it improves biological fidelity and prediction accuracy, offering a new paradigm for single-cell perturbation modeling.

\appendix

\section{Acknowledgments}
This work was supported by National Science and Technology Major Project (No. 2022ZD0115101), National Natural Science Foundation of China Project (No. 623B2086), National Natural Science Foundation of China Project (No. U21A20427), Project (No. WU2022A009) from the Center of Synthetic Biology and Integrated Bioengineering of Westlake University, and the Zhejiang Province Selected Funding for Postdoctoral Research Projects (ZJ2025113).

\bibliography{aaai2026}

\end{document}